\pdfoutput=1

\documentclass[11pt]{article}

\usepackage{naacl2021}

\usepackage{times}
\usepackage{latexsym}

\usepackage{graphicx}
\usepackage{xcolor}
\usepackage{ifthen}

\newcommand{\uc}[1]{\textcolor{purple}{#1}}
\newcommand{\ct}[1]{\textcolor{red}{#1}}
\newcommand{\cc}[1]{\textcolor{red}{#1}}
\newcommand{\yy}[1]{\textcolor{blue}{#1}}

\newcommand{\cy}[1]{\textcolor{orange}{#1}}

\newcommand{\comment}[1]{}

\newcommand{\switch}[1]{%
   \ifthenelse{\equal{#1}{0}}{\renewcommand{\ct}[1]{}}{}}
\switch{0}
\newcommand{\switchyy}[1]{%
   \ifthenelse{\equal{#1}{0}}{\renewcommand{\yy}[1]{}}{}}
\switchyy{1}
\newcommand{\switchcy}[1]{
   \ifthenelse{\equal{#1}{0}}{\renewcommand{\cy}[1]{}}{}}
\switchcy{1}
\usepackage{amsmath,amssymb,amsfonts}

\usepackage[T1]{fontenc}

\usepackage[utf8]{inputenc}

\usepackage{microtype}

\title{ZS-BERT: Towards Zero-Shot Relation Extraction with Attribute Representation Learning}

\author{Chih-Yao Chen \\
  Progam of Data Science \\
  National Taiwan University \\
  Taipei, Taiwan \\
  \texttt{r09946001@ntu.edu.tw} \\\And
  Cheng-Te Li \\
  Institute of Data Science \\
  National Cheng Kung University \\
  Tainan, Taiwan \\
  \texttt{chengte@ncku.edu.tw} \\}

\begin{document}
\maketitle

\begin{abstract}
While relation extraction is an essential task in knowledge acquisition and representation, and new-generated relations are common in the real world, less effort is made to predict unseen relations that cannot be observed at the training stage. 
In this paper, we formulate the zero-shot relation extraction problem by incorporating the text description of seen and unseen relations. We propose a novel multi-task learning model, zero-shot BERT (ZS-BERT), to directly predict unseen relations without hand-crafted attribute labeling and multiple pairwise classifications.
Given training instances consisting of input sentences and the descriptions of their relations, ZS-BERT learns two functions that project sentences and relation descriptions into an embedding space by jointly minimizing the distances between them and classifying seen relations. By generating the embeddings of unseen relations and new-coming sentences based on such two functions, we use nearest neighbor search to obtain the prediction of unseen relations.
Experiments conducted on two well-known datasets exhibit that ZS-BERT can outperform existing methods
by at least 13.54\% improvement on F1 score.


\end{abstract}

\section{Introduction}
\label{sec-intro}

Relation extraction is an important task in the natural language processing field, which aims to infer the semantic relation between a pair of entities within a given sentence. There are many applications based on relation extraction, such as extending knowledge bases (KB)~\cite{lin2015learning} and improving question answering task~\cite{xu-etal-2016-question}. Existing approaches to this task usually require large-scale labeled data. However, the labeling cost is a considerable difficulty. Some recent studies generate labeled data based on distant supervision~\cite{mintz-etal-2009-distant,ji2017distant}.
Nevertheless, when putting the relation extraction task in the wild,
existing supervised models cannot well 
recognize the relations of instances that are extremely rare or even never covered by the training data. 
That said, in the real-world setting, 
we should not presume the relations/classes of new-coming sentences are always included in the training data. Thus it is crucial to invent new models to predict \textit{new classes} that are not defined or observed beforehand. Such a task is referred as \textit{zero-shot learning} (ZSL)~\cite{norouzi2013zeroshot, lampert2014attr, ba2015predicting, kodirov2017semantic}.
The idea of ZSL is to connect seen and the unseen classes by finding an intermediate semantic representation. Unlike the common way to train a supervised model, seen and unseen classes are disjoint at training and testing stages. Hence, ZSL models need to generate transferable knowledge between them. With a model for ZSL relation extraction, we will be allowed to extract unobserved relations, and to deal with new relations resulting from the birth of new entities. 
Existing studies on ZSL relation extraction are few and face some challenges. First, while the typical study~\cite{levy-etal-2017-zero} cannot perform zero-shot relation classification without putting more human effort on it, as they solve this problem via pre-defining question templates. However, it is infeasible and impractical to manually create templates of new-coming unseen relations under the zero-shot setting. We would expect a model that can produce accurate zero-shot prediction without the effort of hand-crafted labeling. In this work, we take advantage of the description of relations, which are usually publicly available, to achieve the goal. Second, although there exists studies that also utilize the accessibility of the relation descriptions~\cite{obamuyide-vlachos-2018-zero}, 
they simply treat zero-shot prediction as the text entailment task and only output a binary label that indicates whether the entities in the input sentence can be depicted by a given relation description. Such problem formulation requires the impractical execution of multiple classifications over all relation descriptions, and cannot make relations comparable with each other.


\begin{figure}
\includegraphics[width=1.0\linewidth]{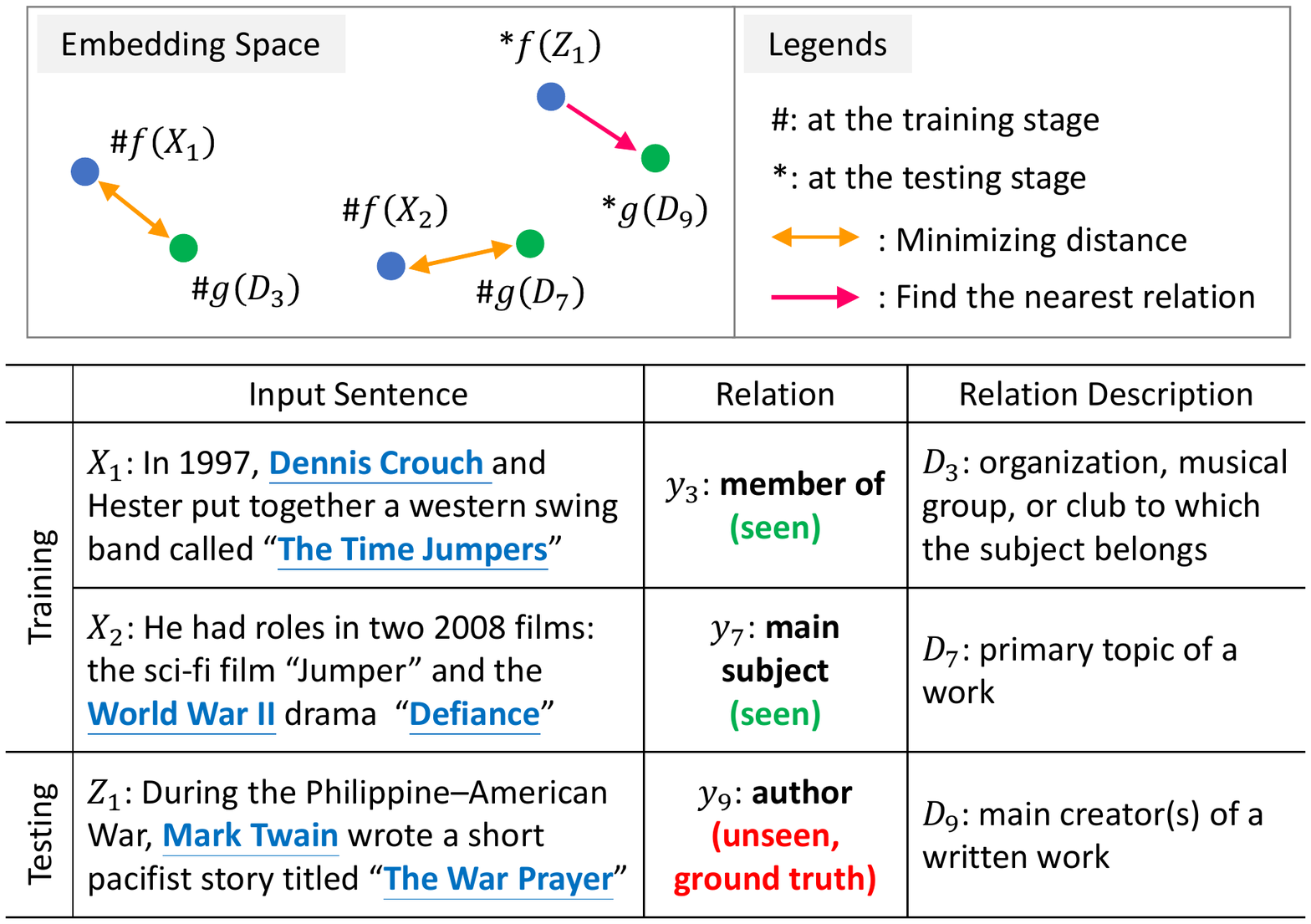}
\caption{An example for elaborating our ZS-BERT.
} 
\label{fig_intro}
\end{figure}

This paper presents a novel model, \textit{Zero-shot BERT} (\textbf{ZS-BERT}), to perform zero-shot learning for relation extraction to cope with the challenges 
mentioned above.
ZS-BERT takes two model inputs. One is the \textit{input sentence} containing the pair of target entities, and the other is the \textit{relation description}, i.e., text describing the relation of two target entities. 
The model output is the \textit{attribute vector}\footnote{The terms, ``attribute vector'', ``embedding'', and ``representation'', are used interchangeably throughout this paper.} depicting the relation. The attribute vector can be considered as a semantic representation of the relation, and will be used to generate the final prediction of unseen relations. 
We think a better utilization of relation descriptions by representation learning is more cost-effective than collecting tons of instances with labeled relations. Therefore, an essential benefit of ZS-BERT is free from heavy-cost crowdsourcing or annotation, i.e., annotating what kind of attribute does a class have, which is commonly used in zero-shot learning problem~\cite{LU2018129,learnobj2009}. 

Figure~\ref{fig_intro} depicts the overview of the proposed ZS-BERT, which consists of five steps. Each training instance is a pair of input sentence $X_i$ and its corresponding relation's description $D_j$.
First, we learn a projection function $f$ that projects the input sentence $X_i$ to its corresponding attribute vector, i.e., sentence embedding.
Second, we learn another mapping function $g$ 
that encodes the relation description $D_j$ as into its corresponding attribute vector, which is the semantic representation of $D_j$.
Third, given the training instance $(X_i,D_j)$,
we train ZS-BERT by minimizing the distance between attribute vectors $f(X_i)$ and $g(D_j)$ in the embedding space. 
Fourth, with the learned $g(D_l)$, we are allowed to project the unseen relation's description $D_l$ into the embedding space so that unseen classes can be as separate as possible for zero-shot prediction. 
Last, given a new input sentence $Z_k$, we can use its attributed vector $f(Z_k)$ to find the nearest neighbor in the embedding space as the final prediction.
In short, the main idea of ZS-BERT is to 
learn the representations of relations based on their descriptions, and to align the representations with input sentences, at the training stage. In addition, we exploit the learned alignment projection functions $f$ and $g$ to generate the prediction of unseen relations for the new sentence so that the zero-shot relation extraction can be achieved. Our contributions can be summarized as below.
\comment{
We have conducted several experiments not only to study the performance of our model with comparison to the others, but also the effect of parameters and case study to show the goodness and limitation of our model. We have shown that while supervised model are still struggling on predicting unseen classes, our proposed model can have good ability to recognize them. 
}

\begin{itemize}
\item Conceptually, we formulate the zero-shot relation extraction problem by leveraging text descriptions of seen and unseen relations. To the best of our knowledge, we are the first attempt to directly predict unseen relation under the zero-shot setting via learning the representations from relation descriptions. 
\item Technically, we propose a novel deep learning-based model, ZS-BERT\footnote{Code and implementation details can be accessed via: \url{https://github.com/dinobby/ZS-BERT}.}, to tackle the zero-shot relation extraction task. ZS-BERT learns the projection functions to align the input sentence with its relation in the embedding space, and thus is capable of predicting relations that were not seen during the training stage.
\item Empirically, experiments conducted on two well-known datasets exhibit that ZS-BERT can significantly outperform state-of-the-art methods for predicting unseen relations under the ZSL setting. 
We also show that ZS-BERT can be quickly adapted and generalized to few-shot learning when a small fraction of labeled data for unseen relations is available.
\end{itemize}

\section{Related Work}
\label{sec-related}

\paragraph{BERT-based Relation Extraction.} Contextual representation of words is effective for NLP tasks.
BERT~\cite{devlin-etal-2019-bert} is a pre-training language model that learns useful contextual word representations.
BERT can be moderately adopted for supervised or few-shot relation extraction.
R-BERT~\cite{wu2019enrich} utilize BERT to generate contextualized word representation, along with entities' information to perform supervised relation extraction and have shown promising result.
BERT-PAIR~\cite{gao-etal-2019-fewrel} makes use of the pre-train BERT sentence classification model for few-shot relation extraction. By pairing each query sentence with all sentences in the support set, they can get the similarity between sentences by pre-trained BERT, and accordingly classify new classes with a handful of instances. 
These models aim to solve the general relation extraction task, which are more or less having ground truth, rather than having it under the zero-shot setting. 


\noindent\textbf{Zero-shot Relation Extraction.}
Relevant studies on zero-shot relation extraction are limited. To the best of our knowledge, there are two most similar papers, which consider zero-shot relation extraction as two different tasks.
\citet{levy-etal-2017-zero} treat zero-shot relation extraction as a question answering task. They manually define 10 question templates to represent relations, and generate the prediction by training a reading comprehension model to answer which relation satisfies the given sentence and question. However, it is required to have human efforts on defining question templates for unseen relations so that ZSL can be performed. Such annotation by domain knowledge is unfeasible in the wild when more unseen relations come. On the contrary, the data requirement of ZS-BERT is relatively lightweight. For each relation, we only need one description that could express the semantic meaning. The descriptions of relations are easier to be collected as we may access them from open resources. Under such circumstances, we may be free from putting additonal effort to the annotation.

\citet{obamuyide-vlachos-2018-zero} formulate ZSL relation extraction as a textual entailment task, which requires the model to predict whether the input sentence containing two entities matches the description of a given relation. 
They use Enhanced Sequential Inference Model (ESIM)~\cite{chen2016enhanced} and Conditioned Inference Model (CIM)~\cite{rocktaschel2015reasoning} as their entailment methods. By pairing each input sentence with every relation description, they train the models to answer whether the paired texts are contradiction or entailment. This allow the model to inference on input sentence and unseen relation description pair, thus is able to predict unseen relation accordingly.




\section{Problem Definition}
\label{sec-problem}

Let ${Y_s} = \{{y^1_s}, ... , {y^n_s}\}$ and ${Y_u} = \{{y^1_u} , ... , {y^m_u}\}$ denote the sets of seen and unseen relation labels, respectively, in which $n=|Y_s|$ and $m=|Y_u|$ are the numbers of relations in two sets. Such two sets are disjoint, i.e., ${Y_s} \cap {Y_u} = \emptyset$. For each relation label in seen and unseen sets, we denote the corresponding attribute vector as ${a^i_s} \in \mathbb{R}^{n\times d}$ and ${a^i_u} \in \mathbb{R}^{m\times d}$, respectively. 
Given the training set with $N$ samples, consisting of input sentence $X_i$, entities ${e_{i1}}$ and ${e_{i2}}$, and the description $D_i$ of the corresponding seen relation $y^j_s$, denoted as $\{S_i = ({X_i}, {e_{i1}}, {e_{i2}}, {D_i}, {y^j_s})\}^N_{i=1}$. 
Our goal is to train a zero-shot relation extraction model $\mathcal{M}$, i.e., $\mathcal{M}(S_i)\rightarrow y^i_s \in Y_s$, based on the training set such that using $\mathcal{M}$ to predict the \textit{unseen} relation $y^k_u$ of a testing instance $S'$, i.e., $\mathcal{M}(S')\rightarrow y^j_u\in Y_u$, can achieve as better as possible performance. 

We train the model $\mathcal{M}$ so that the semantics between input sentence and relation description can be aligned. We learn $\mathcal{M}$ by minimizing the distance between two embedding vectors $f(X_i)$ and $g(D_i)$, where learnable functions $f$ and $g$ project $X_i$ and $D_i$ into the embedding space, respectively.
When new unseen relation $y^j_u$ and its description is in hand, we can project the description of $y^j_u$ to the embedding space by function $g$.
When testing, new instance $S'=({Z_j}, {e_{j1}}, {e_{j2}}, {D_j})$ is input, in which $Z_i$ denotes new sentence containing entities ${e_{j1}}$ and ${e_{j2}}$, we project $Z_i$ to the embedding space by our learned function $f$, and find the nearest neighboring unseen relation ${y^j_u}$, where ${Z_i}$ and ${y^i_u}$ are both unknown at the training stage.
\section{The Proposed ZS-BERT Model}
\label{sec-model}

\begin{figure}
   \includegraphics[width=\linewidth]{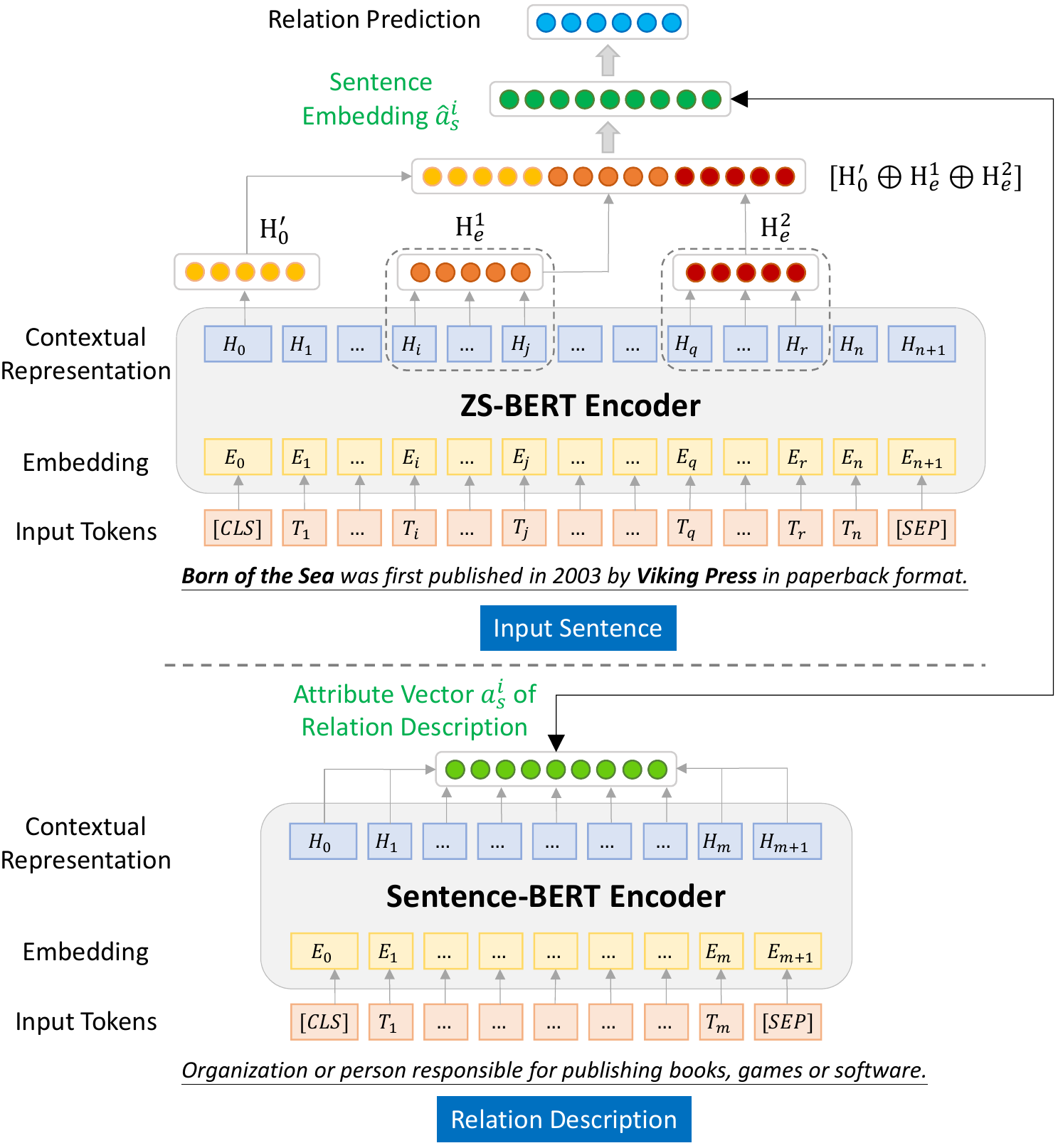}
   \caption{The overall architecture of our model.}
   \label{fig_model}
\end{figure}

We give an overview of our ZS-BERT in Figure~\ref{fig_model}.
The input sentence $X_i$ is tokenized and sent into the upper-part ZS-BERT encoder to obtain contextual representation. We specifically extract the representation of [CLS], $H_0$, and two entities' representations $H_e^1,H_e^2$, and then concatenate them to derive sentence embeddings $\hat{a}_s^i$, by a fully-connected layer and activation operation. In the bottom part, we use Sentence-BERT~\cite{reimers-gurevych-2019-sentence} to obtain attribute vector $a_s^i$ for seen relations by encoding the corresponding description of relation $D_i$. We train ZS-BERT under a multi-task learning structure. One task is to minimize the distance between attribute vector $a_s^i$ and sentence embedding $\hat{a}_s^i$. 
The other is to classify the seen relation $y^j_s$ at the training stage, in which 
a softmax layer that accepts relation embedding is used to produce the relation classification probability. At the testing stage, by obtaining the embeddings of new-coming sentences and unseen relations, we use $\hat{a}_s^i$ and nearest neighbor search to obtain the prediction of unseen relations. 

\subsection{Learning Relation Attribute Vectors}
For each seen and unseen relation, we learn its representation that depicts the corresponding semantic attributes based on relation description $D_i$. Most relations are well-defined and their descriptions are accessible from online open resources such as Wikidata~\footnote{\url{https://www.wikidata.org}}. We feed relation description $D_i$ into a pre-trained Sentence-BERT encoder~\cite{reimers-gurevych-2019-sentence} to generate the sentence-level representation as the attribute vector $a^i$ of relations. This procedure is shown in the bottom part of Figure~\ref{fig_model}. The ground truth relation of the example is \textit{publisher}, along with its description \textit{Organization or person responsible for publishing books, games or software.} We feed only the relation description to the Sentence-BERT in order to get the attribute vector.
That said, we consider the derived Sentence-BERT to be a projection function $g$ that transforms the relation description $D_i$ into $a^i$. 
Note that the relation attribute vectors produced by Sentence-BERT are fixed during model training.


\subsection{Input Sentence Encoder}
We utilize BERT~\cite{devlin-etal-2019-bert} to generate the contextual representation of each token. We first tokenize the input sentences $X_i$ with WordPiece tokenization~\cite{sennrich-etal-2016-neural}. 
Two special tokens [CLS] and [SEP] are appended to the first and last positions, respectively. Since the entity itself does matter in relation extraction, 
we use an entity marker vector, consisting of all zeros except the indices that entities appear in a sentence,
to indicate the positions of entities $e_{i1}$ and $e_{i2}$. 
Let $H_0$ be the hidden state of the first special token [CLS].
We use a $tanh$ activation function, together with a fully connected layer, to derive the representation vector $H'_0$, given by:
    $H'_0 = {W_0}[{tanh({H_0})}]+{b_0}$,
where $W_0$ and $b_0$ are learnable parameters for weights and biases. We obtain the hidden state vectors of two entities, $H_e^1$ and $H_e^2$, by averaging their respective tokens' hidden state vectors.
The entity can be recognized via simple element-wise multiplication between entity marker vector and token hidden vector.
Specifically, if an entity $e$ consists of multiple tokens and the indices range from $q$ to $r$, we average the hidden state vectors, and also add an activation operation with a fully connected layer to generate its representation of that entity, given by:
${{H^c_e} = {W_e}{\left[tanh\left(\frac{1}{r-q+1}\sum^r_{t=q}{H_t}\right)\right]+b_e}}$,
where $c=1,2$.
Note that the representations of two entities $H^c_e (c=1,2)$ in the sentence shares the same parameters $W_e$ and $b_e$.
Then we learn the attribute vector $\hat{a}^i_s$ by concatenating $H'_0$, $H_e^1$, and $H_e^2$, followed by a hidden layer, given by:
\begin{equation}
\hat{a}^i_s = W_1(tanh([{H'_0} \oplus {H_e^1} \oplus{H_e^2}])) + b_1,
\end{equation}
where $W_1$ and $b_1$ are learnable parameters 
, the dimensionality of $\hat{a}^i_s$ is $d$,
and $\oplus$ is the concatenation operator.

\begin{table*}
\centering
\caption{Datasets. ``avg. len.'' is average sentence len.}
\label{tab_dataset}
\begin{tabular}{ccccl} 
\hline
         & \multicolumn{1}{l}{\#instances} & \multicolumn{1}{l}{\#entities} & \multicolumn{1}{l}{\#relations} & avg. len.  \\ 
\hline
Wiki-KB & 1,518,444                           & 306,720                            & 354                                &23.82                       \\
Wiki-ZSL & 94,383                           & 77,623                            & 113                                &24.85                 \\
FewRel   & 56,000                              & 72,954                             & 80                                 &24.95                       \\
\hline
\end{tabular}
\end{table*}

\subsection{Model Training}
The training of our ZS-BERT model consists of two objectives. The first is to 
minimize the distance 
between input sentence embedding $a_s^i$ and the corresponding relation attribute vector $\hat{a}_s^i$ (i.e., \textit{positive} pairs), meanwhile to ensure embedding pairs between input sentence embedding and mis-matched relation (i.e., \textit{negative} pairs)
to be farther away from each other. The black arrow connecting $a_s^i$ and $\hat{a}_s^i$ in Figure~\ref{fig_model} is a visualization to indicate that we take both $a_s^i$ and $\hat{a}_s^i$ into consideration to achieve this goal. This is also reflected in the first term of our proposed loss function introduced below. 
The second objective is to maximize the accuracy of relation classification based on seen relations using cross entropy loss.
We transform the relation embedding, along with a softmax layer, to generate a $n$-dimensional ($n = |Y_s|$) classification probability distribution over seen relations:
$p(y_s|X_i,\theta) = softmax(W^*(tanh(\hat{a}^i_s)) + b^*)$,
where $y_s \in Y_s$ is the seen relation, $\theta$ is the model parameter, ${W^*} \in \mathbb{R}^{n\times{h}}$, $h$ is the dimension of hidden layer, and ${b^*} \in \mathbb{R}^{n}$. Note that we do not use the probability distribution but the input sentence embedding $\hat{a}^i_s$ produced \textit{intermediately} for predicting unseen relations under zero-shot settings. 

The objective function of ZS-BERT is as follows:
\begin{equation}
\label{lossfunc}
\begin{aligned}
L &=  (1 - \alpha) \sum^N_i{\max(0, \gamma - a^i_s \cdot \hat{a}^i_s + \max_{i\neq j}{(a^i_s \cdot \hat{a}^j_s)})}\\
&- \alpha \sum^N_i{y^i_slog(\hat{y^i_s})},
\end{aligned}
\end{equation}
where $N$ is the number of samples, $a^i_s$
is the relation attribute vector,
and $\hat{a}^i_s$ is
the input sentence embedding. The first term in Eq.~(\ref{lossfunc}) sets a margin $\gamma > 0$ such that the inner product of the positive pair (i.e., $a^i_s \cdot \hat{a}^i_s$) must be higher than the maximum of the negative one (i.e., $\max_{i\neq j}{(a^i_s \cdot \hat{a}^j_s)}$) for more than a pre-decided threshold $\gamma$.
With the introduction of $\gamma$, the loss will be increased owing to the difference between the positive and the closest negative pairs. This design of loss function can be viewed as ranking the correct relation attribute higher than the closest incorrect one. In addition, $\gamma$
is also utilized to avoid the embedding space from collapsing. If we consider only minimizing the distance of positive pair 
using loss like Mean Squared Error, the optimization may lead to the result that every vector in the embedding space is too close to one another. We will examine how different $\gamma$ values affect the performance in the experiment.
To maintain low computational complexity, we consider only those mismatched relations within a batch as the negative samples $j$.
The second term in Eq.~(\ref{lossfunc}) is a commonly used cross entropy loss, which decreases as the prediction $\hat{y}^i_s$ is correctly classified. Such a multi-task structure is expected to refine the input sentence embeddings and simultaneously bring high prediction accuracy of seen relations.

\subsection{Generating Zero-Shot Prediction}
With the trained model, when the descriptions of new relations are in hand, we can generate their attribute vectors $a^j_u$.
As the new input sentence $Z_i$ arrives, we can also produce its sentence embedding $\hat{a}^i_u$ via:
$\hat{a}^i_u = W_1(tanh([H'_0 \oplus H_e^1 \oplus H_e^2])) + b_1$,
where $W_1$ and $b_1$ are learned parameters.
The prediction on unseen relations can be achieved by the nearest neighbor search. For the input sentence embedding $\hat{a}^i_u$, we find the nearest attribute vector $a^j_u$ and consider the corresponding relation as the predicted unseen relation. This can be depicted by:
$C(Z_i) = \operatorname*{argmin}_j dist(\hat{a}_u^i, a_u^j)$,
where function $C$ returns the predicted relation of new input sentence $Z_i$, $a_u^j$ is the $j$-th attribute vector among all unseen relations in the embedding space, $\hat{a}_u^i$ is the new input sentence embedding,
and $dist$ is a distance computing function. Here negative inner product is used as $dist$ since we aim to consider the nearest neighboring relation as the predicted outcome. 
\section{Experiments}
\label{sec-experiment}

\comment{
We conduct experiments to answer the several evaluation questions:
    (1) Can ZS-BERT outperform existing zero-shot relation extraction models? By what percentage of improvement can it make?
    (2) How does different numbers of unseen relations affect performance when doing zero-shot learning?
    (3) To what extent is ZS-BERT sensitive to hyperparameters?
    (4) Can ZS-BERT be generalized to few-shot learning and how well can it carry out prediction when there is a little supervision available?
    (5) What are the strengths and limitations of ZS-BERT via case studies? 
}

\subsection{Evaluation Settings}
\textbf{Datasets.} Two datasets are employed, \textbf{Wiki-ZSL} and \textbf{FewRel}~\cite{han-etal-2018-fewrel}.
Wiki-ZSL is originated from Wiki-KB~\cite{sorokin-gurevych-2017-context}, and is generated with distant supervision. That said, in Wiki-ZSL, \textit{entities} are extracted from complete articles in Wikipedia, and are linked to the Wikidata knowledge base so that their \textit{relations} can be obtained. Since $395,976$ instances (about $26\%$ of the total data) do not contain relations in the original Wiki-KB data, we neglect
instances with relation ``none''. To ensure having sufficient data instances for each relation in zero-shot learning, we further filter out the relations that appear fewer than 300 times. Eventually, we can have yields \textbf{Wiki-ZSL}, a subset of Wiki-KB.

On the other hand, FewRel~\cite{han-etal-2018-fewrel} is compiled by a similar way to collect entity-relation triplet with sentences, but had been further filtered by crowd workers. This ensures the data quality and class balance.
Although FewRel is originally proposed for few-shot learning, it is also suitable for zero-shot learning as long as the relation labels within training and testing data are disjoint. 
The statistics of Wiki-KB, Wiki-ZSL and FewRel datasets are shown in Table~\ref{tab_dataset}.

\textbf{ZSL Settings.} We randomly select $m$ relations as \textit{unseen} ones ($m = |Y_u|$), and randomly split the whole dataset into training and testing data, meanwhile ensuring that these $m$ relations do not appear in training data so that ${Y_s} \cap {Y_u} = \emptyset$.
We repeat the experiment $5$ times for random selection of $m$ relations and random training-testing splitting, and report the average results. We will also vary $m$ to examine how performance is affected. We use \textit{Precision} (P), \textit{Recall} (R), and \textit{F1} as the evaluation metrics. 
As for the hyperparameters and configuration of ZS-BERT, we use Adam~\cite{kingma2014adam} as the optimizer, in which the initial learning rate is $5e-6$, the hidden layer size is $768$, the dimension of input sentence embedding and attribute vector is $1024$, the batch size is $4$, $\gamma=7.5$, and $\alpha=0.4$. 

\textbf{Competing Methods.} 
The compared methods consist of two categories, supervised relation extraction (SRE) models and text entailment models. The former includes CNN-based SRE~\cite{zeng-etal-2014-relation}, Bi-LSTM SRE~\cite{zhang-etal-2015-bidirectional}, Attentional Bi-LSTM SRE~\cite{zhou-etal-2016-attention}, and R-BERT~\cite{wu2019enrich}.
These SRE models
use different ways to extract features from the input sentences and perform prediction. They have achieved great performance with fully supervision but fail to carry out zero-shot prediction. To make them capable of zero-shot prediction, also to have fair comparison, instead of originally using a softmax layer to output a probability vector whose dimension is equal to the seen relations, we change the last hidden layer of each SRE competing method to a fully-connected layer with a $tanh$ activation function, and the embedding dimension $d$ is the same as ZS-BERT. The nearest neighbor search is applied over input sentence embeddings and relation attribute vectors to generate zero-shot prediction.

Two text entailment models, ESIM~\cite{chen2016enhanced} and CIM~\cite{rocktaschel2015reasoning}, are also used for comparison.
These two models follow a well-known implementation~\cite{obamuyide-vlachos-2018-zero} that formulates zero-shot relation extraction as a text entailment task, which accepts sentence and relation description as input, and output a binary label indicating whether they are semantically matched. ESIM uses bi-LSTM~\cite{hochreiter1997long,graves2005framewise} to encode two input sequences, passes them through the local inference model, and produces the prediction via a softmax layer.
CIM replaces the bi-LSTM block with a conditional version, i.e., 
the representation of sentence is conditioned on its relation description. 
Note that although there exist other zero-shot relation extraction approaches such as the approach proposed by \citet{levy-etal-2017-zero}, their approach to formulate the ZSL task and their data requirement are quite different with our present work. To be specific, their method requires pre-defined question template, whereas our model does not. Hence it would be unfair to compare with those approaches. 

\subsection{Experimental Results}
\begin{table}[]
\centering
\small
\caption{Results with different $m$ values in percentage.}
\label{tab_performance}
\resizebox{\linewidth}{!}{%
\begin{tabular}{lcccccc}
\hline
\multicolumn{1}{c}{\textbf{}} &
  \multicolumn{3}{c}{Wiki-ZSL} &
  \multicolumn{3}{c}{FewRel} \\ \hline
 &
  \multicolumn{3}{c}{m=5} &
  \multicolumn{3}{c}{m=5} \\
\multicolumn{1}{c}{} &
  P &
  R &
  F1 &
  P &
  R &
  F1 \\ \hline
CNN &
  30.31 &
  32.17 &
  30.92 &
  36.41 &
  38.69 &
  37.42 \\
Bi-LSTM &
  36.73 &
  40.44 &
  38.62 &
  41.99 &
  50.25 &
  45.66 \\
Att Bi-LSTM &
  35.58 &
  41.26 &
  38.21 &
  39.52 &
  47.24 &
  42.95 \\
R-BERT &
  39.22 &
  43.27 &
  41.15 &
  42.19 &
  48.61 &
  45.17 \\
ESIM &
  \multicolumn{1}{l}{48.58} &
  \multicolumn{1}{l}{47.74} &
  \multicolumn{1}{l}{48.16} &
  \multicolumn{1}{l}{56.27} &
  \multicolumn{1}{l}{58.44} &
  \multicolumn{1}{l}{57.33} \\
CIM &
  \multicolumn{1}{l}{49.63} &
  \multicolumn{1}{l}{48.81} &
  \multicolumn{1}{l}{49.22} &
  \multicolumn{1}{l}{58.05} &
  \multicolumn{1}{l}{61.92} &
  \multicolumn{1}{l}{59.92} \\ \hline
\textbf{ZS-BERT} &
  \textbf{71.54} &
  \textbf{72.39} &
  \textbf{71.96} &
  \textbf{76.96} &
  \textbf{78.86} &
  \textbf{77.90} \\ \hline
 &
  \multicolumn{3}{c}{m=10} &
  \multicolumn{3}{c}{m=10} \\
 &
  P &
  R &
  F1 &
  P &
  R &
  F1 \\ \hline
CNN &
  20.86 &
  23.61 &
  22.08 &
  22.37 &
  28.15 &
  24.85 \\
Bi-LSTM &
  25.33 &
  27.91 &
  26.56 &
  24.52 &
  32.02 &
  27.77 \\
Att Bi-LSTM &
  24.98 &
  29.13 &
  26.90 &
  24.24 &
  31.32 &
  27.28 \\
R-BERT &
  26.18 &
  29.69 &
  27.82 &
  25.52 &
  33.02 &
  28.20 \\
ESIM &
  \multicolumn{1}{l}{44.12} &
  \multicolumn{1}{l}{45.46} &
  \multicolumn{1}{l}{44.78} &
  \multicolumn{1}{l}{42.89} &
  \multicolumn{1}{l}{44.17} &
  \multicolumn{1}{l}{43.52} \\
CIM &
  \multicolumn{1}{l}{46.54} &
  \multicolumn{1}{l}{47.90} &
  \multicolumn{1}{l}{45.57} &
  \multicolumn{1}{l}{47.39} &
  \multicolumn{1}{l}{49.11} &
  \multicolumn{1}{l}{48.23} \\ \hline
\textbf{ZS-BERT} &
  \textbf{60.51} &
  \textbf{60.98} &
  \textbf{60.74} &
  \textbf{56.92} &
  \textbf{57.59} &
  \textbf{57.25} \\ \hline
 &
  \multicolumn{3}{c}{m=15} &
  \multicolumn{3}{c}{m=15} \\
 &
  P &
  R &
  F1 &
  P &
  R &
  F1 \\ \hline
CNN &
  14.58 &
  17.68 &
  15.92 &
  14.17 &
  20.26 &
  16.67 \\
Bi-LSTM &
  16.25 &
  18.94 &
  17.49 &
  16.83 &
  27.62 &
  20.92 \\
Att Bi-LSTM &
  16.93 &
  18.54 &
  17.70 &
  16.48 &
  26.36 &
  20.28 \\
R-BERT &
  17.31 &
  18.82 &
  18.03 &
  16.95 &
  19.37 &
  18.08 \\
ESIM &
  \multicolumn{1}{l}{27.31} &
  \multicolumn{1}{l}{29.62} &
  \multicolumn{1}{l}{28.42} &
  \multicolumn{1}{l}{29.15} &
  \multicolumn{1}{l}{31.59} &
  \multicolumn{1}{l}{30.32} \\
CIM &
  \multicolumn{1}{l}{29.17} &
  \multicolumn{1}{l}{30.58} &
  \multicolumn{1}{l}{29.86} &
  \multicolumn{1}{l}{31.83} &
  \multicolumn{1}{l}{33.06} &
  \multicolumn{1}{l}{32.43} \\ \hline
\textbf{ZS-BERT} &
  \textbf{34.12} &
  \textbf{34.38} &
  \textbf{34.25} &
  \textbf{35.54} &
  \textbf{38.19} &
  \textbf{36.82} \\ \hline
\end{tabular}}
\end{table}
\textbf{Main Results.} The experiment results by varying $m$ unseen relations
are shown in Table \ref{tab_performance}. 
First, it can be apparently found that the proposed ZS-BERT steadily outperforms the competing methods over two datasets when targeting at different numbers of unseen relations. The superiority of ZS-BERT gets more significant on $m=5$. Such results not only validate the effectiveness of leveraging relation descriptions, but also prove the usefulness of the proposed multi-task learning structure that better encodes the semantics of input sentences and have relation attribute vectors been differentiated from each other.
Second, although the text entailment models ESIM and CIM perform well among competing methods, their performance is still obviously lower than ZS-BERT. The reason is that their approaches cannot precisely distinguish the semantics of input sentences and relation descriptions in the embedding space. 
Third, we also find that the improvement of ZS-BERT gets larger when $m$ is smaller. Increasing $m$ weakens the superiority of ZS-BERT. 
It is straightforward that as the number of unseen relations increases, it becomes more difficult to predict the right relation since the possible choices have increased. 
We also speculate another underlying reason is that although ZS-BERT can effectively capture the latent attributes for each relation, relations themselves could be to some extent semantically similar to one another, and more unseen relations will increase the possibility that obtains a predicted relation that is semantically close but actually wrong. To verify this conjecture, we will give an example in the case study. 

\begin{figure}[!t]
   \makebox[\linewidth][c]{\includegraphics[width=\linewidth]{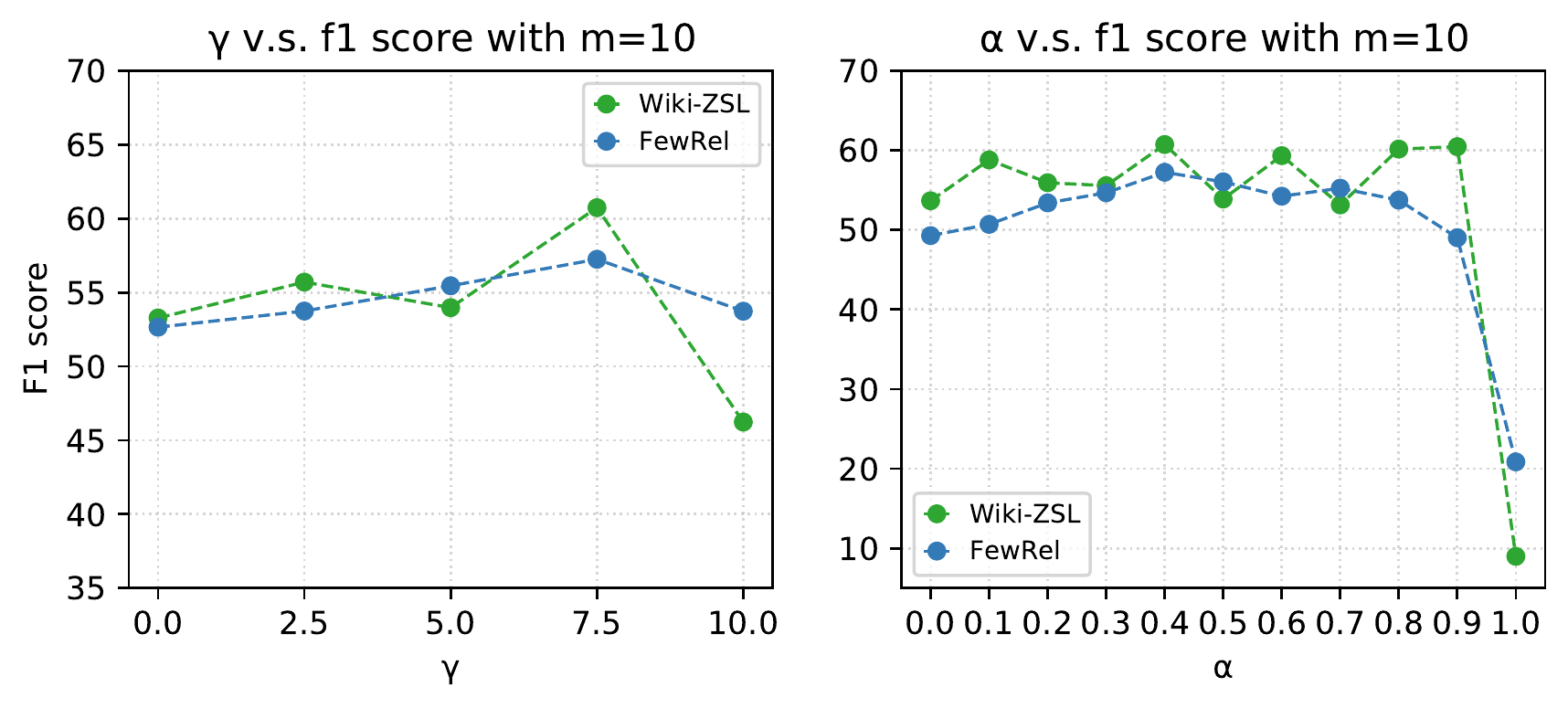}}
   \caption{Effects on varying the margin parameter $\gamma$ and balance coefficient $\alpha$ with $m$=10 on both datasets.
   } \label{fig_gamma_alpha}
\end{figure}

\begin{figure}[!t]
   \makebox[\linewidth][c]{\includegraphics[width=\linewidth]{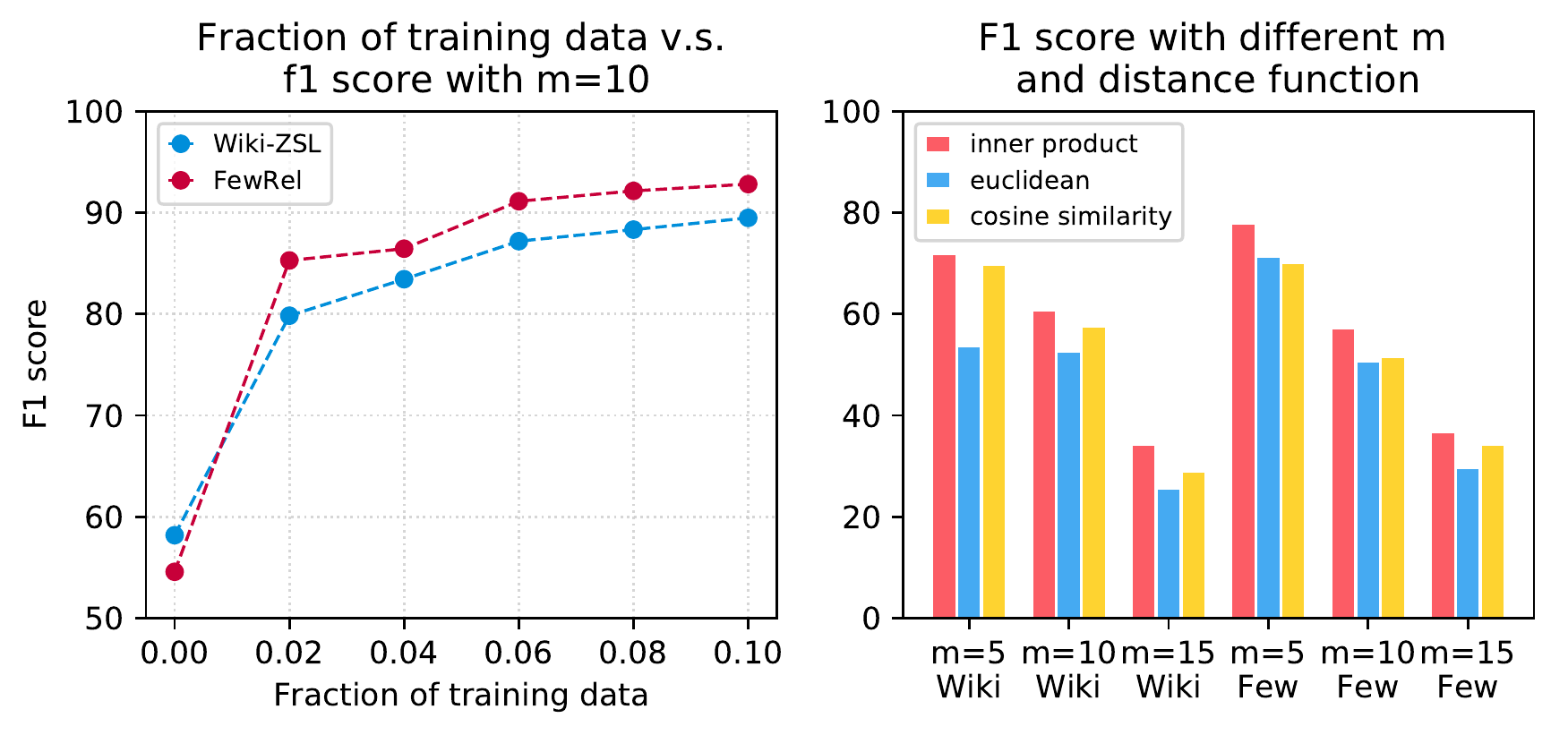}}
   \caption{Left: Results of ZS-BERT with different fractions of unseen instances availble for training, in which $0.0$ refers to the zero-shot result. Right: Results on different distance functions and varied $m$.} \label{fig_fraction}
\end{figure}

\textbf{Hyperparameter Sensitivity.} We examine how primary hyperparameters, including the value of margin parameter $\gamma$ and the balance coefficient $\alpha$ in Eq.~\ref{lossfunc}, affect the performance of ZS-BERT. By fixing $m=10$ and varying $\gamma$ and $\alpha$, the results in terms of F1 scores on two datasets are exhibited on Figure~\ref{fig_gamma_alpha}. 
It is noteworthy that $\gamma$ does have an impact on performance, since it brings the condition on whether to increase the loss value, which is determined by the difference between the positive pair and negative pair. Nevertheless, not always the higher values of $\gamma$ lead to better performance. 
This is reasonable that when $\gamma$ is too low, the distance between the positive pair and negative pair would not be far enough. Thus, when performing nearest neighbor search, it is more likely to reach the wrong relations. In contrast, when $\gamma$ gets too high, it is hard for the training process to converge at the point that the distance between relations is expected to be that high.
We would suggest setting $\gamma=7.5$ to derive satisfying results across datasets.
As for the balance coefficient $\alpha$ in the loss function, we find that $\alpha=0.4$ can achieve the best performance, indicating that the margin loss plays a more significant role in training ZS-BERT. Also notice that when $\alpha=1.0$, the performance drops dramatically, showing that the margin loss is essential to our model. This is also reasonable that since our model relies on the quality of embeddings, therefore totally relying on cross entropy loss leads to failure of zero-shot prediction. The better separation between embeddings of different relations, the more likely our model can generate the accurate zero-shot prediction.
In addition, while the nearest neighbor search is performed to generate the zero-shot prediction, we think the choice of distance computing function $dist()$ can also be an hyperparameter. By applying inner product, Euclidean distance, and the cosine similarity as $dist()$ in ZS-BERT, we report their F1 scores with different $m$ on two datasets in the right of Figure~\ref{fig_fraction}. The results inform us that inner product is a proper distance function for zero-shot relation extraction with ZS-BERT.


\begin{table}[!t]
\centering
\small
\caption{List of four cases, in which head and tail entities are highlighted by green and blue, respectively.}
\label{tab_case_study}
\resizebox{\linewidth}{!}{%
\begin{tabular}{|l|l|c|c|}
\hline
  & \multicolumn{1}{c|}{Input Sentence} & True & Predicted  \\
\hline
(1) & \begin{tabular}[c]{@{}l@{}}~When promoting \textbf{\textcolor[rgb]{0,0.502,0}{Anaconda}}, \textbf{Minaj} confirmed \\\ plans of a tour in support of \textbf{\textcolor[rgb]{0,0.365,1}{The Pinkprint}} in \\\ an interview with Carson Daly on AMP Radio.\end{tabular}                                                                      & tracklist     & publisher           \\ 
\hline
(2) & \begin{tabular}[c]{@{}l@{}}\textbf{~\textcolor[rgb]{0,0.502,0}{Heaven}} and \textcolor[rgb]{0,0.365,1}{\textbf{Hell} }as they are understood in Christian \\
\ theology are roughly analogous to the Jewish Olam \\\ habah and Gehenna , with certain major differences.\end{tabular} & \begin{tabular}[c]{@{}c@{}}opposite\\of \end{tabular} & \begin{tabular}[c]{@{}c@{}}influenced\\by \end{tabular} \\
\hline
(3) & \begin{tabular}[c]{@{}l@{}}~(TOEO) was an MMORPG set in the world of the \\\  popular \textbf{\textcolor[rgb]{0,0.502,0}{Namco}~}PlayStation title, \textbf{\textbf{~\textcolor[rgb]{0,0.365,1}{Tales of Eternia}}. }\end{tabular}                                                                                                                            & publisher     & manufacturer        \\ 
\hline
(4) & \begin{tabular}[c]{@{}l@{}}~The inverse of \textbf{\textcolor[rgb]{0,0.502,0}{admittance }}is \textcolor[rgb]{0,0.365,1}{\textbf{impedance}},\\\ and the real part of admittance is conductance.\end{tabular}& \begin{tabular}[c]{@{}c@{}}opposite\\of \end{tabular} & \begin{tabular}[c]{@{}c@{}}influenced\\by \end{tabular} \\
\hline
\end{tabular}}
\end{table}
\begin{figure*}
\centering
   \includegraphics[width=\textwidth]{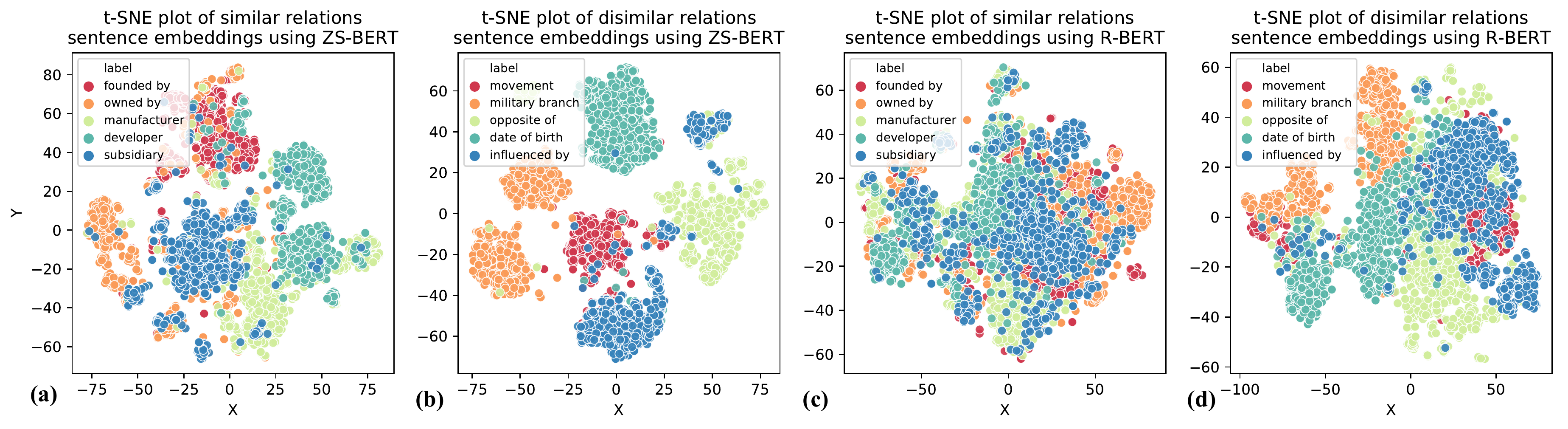}
   \caption{t-SNE visualization of the sentence embeddings for similar (a)(c) and dissimilar relations (b)(d).} \label{fig_tsne}
\end{figure*}

\textbf{Few-shot Prediction.} To understand the capability of ZS-BERT, we conduct the experiment of few-shot prediction. By following the setting of an existing work~\cite{obamuyide-vlachos-2018-zero}, we make a small fraction of unseen data instances available at the training stage. That said, for each originally unseen relation, we move a small fraction of its sentences, along with the relation description, from the testing to the training stage. By varying the fraction in x-axis, we report the results of few-shot prediction in Figure~\ref{fig_fraction}. We can find that that ZS-BERT can reach about 80\% on F1 score with only 2\% of unseen instances as supervision. Such results demonstrate the ability to recognize rare samples and the capability of few-shot learning for the proposed ZS-BERT. As expected, the more instances belonging to unseen relations available at the training stage, the higher the F1 score is. When the fraction equals to 10\%, ZS-BERT can even achieve 90\% F1 score on Wiki-ZSL dataset. 






\subsection{Case Study}
We categorize four types of incorrectly predicted unseen relations for the analysis:
(1) The predicted relation is not precise for the targeted entity pair but may be suitable for other entities that also appear in the sentence. 
(2) The true relation is not appropriate because it comes from distant supervision. 
(3) The predicted relation is ambiguous or is a synonym of other relations. 
(4) The relation is wrongly predicted but should be able to be correctly classified. 
For each of these four types, we provide an example listed in Table~\ref{tab_case_study}.
In case (1), the targeted entities are \textbf{Anaconda} and \textbf{The Pinkprint}, and ZS-BERT yields \textit{publisher} as the prediction, which is actually correct if the targeted entities are \textbf{Anaconda} and \textbf{Minaj}. This shows ZS-BERT is able to infer the possible relation for entities in the given sentence, but sometimes could be misled by non-targeted entities even though we have an entity mask to indicate the targeted entities.
In case (2), it shows the noise originated from distant labeling. That is, even human being cannot identify the relation between \textbf{Heaven} and \textbf{Hell} is \textit{opposite of} in this specific sentence. They just happened to appear together and their relation recorded in Wikidata is \textit{opposite of}. 
In case (3), the predicted unseen relation is \textit{manufacturer}, while the ground truth is \textit{publisher}. Both \textit{manufacturer} and \textit{publisher} describe someone make or produce something, although their domains are slightly different. This exhibits the capability of ZS-BERT to identify the input sentence with an \textit{abstract} attribute because relations possessing similar semantics will have similar attribute vectors in the embedding space. 
Finally, in case (4), the model gives a wrong prediction that is not even close or related, which may due to the noise or information loss when transferring knowledge between relations.

Among these four groups, we are especially interested in case (3) since the semantic similarity between relations in the embedding space greatly impacts the performance. We select five semantically-distant relations, and the other five relations that possess similar semantics between two or three of them, to inspect their distributions in the embedding space. We feed sentences with these relations and generate their embeddings using ZS-BERT and R-BERT~\cite{wu2019enrich} for comparison. We choose R-BERT because it is the strongest embedding-based competing method for zero-shot prediction by nearest neighbor search. Note that since the predictions by text entailment-based models, ESIM and CIM, neither resort to similarity search nor directly predict unseen relation at one time, we cannot have them compared in this analysis. We visualize the embedding space by t-SNE~\cite{maaten2008visualizing}, as shown in Figure~\ref{fig_tsne}. We can find that when the relations are somewhat similar in their meanings (Figure~\ref{fig_tsne}(a),(c)), some of the data points are mingled with different clusters, as they indeed have close semantic relationships. Take \textit{subsidiary} and \textit{owned by} as examples, \textit{Company A is a subsidiary of company B} and \textit{Company A is owned by company B} refer to the same thing. This happens on both ZS-BERT and R-BERT but to a different extent. It is obvious that the embeddings produced by R-BERT are more tangled. We also plot the other five relations that there is no ambiguity among them (Figure~\ref{fig_tsne}(b),(d)). Apparently their embeddings are more separated between different relations. It is also obvious that the embeddings generated by ZS-BERT lead to larger inter-relation distance. This again exhibits the usefulness of the proposed ranking loss and multi-task learning structure.  

\section{Conclusions}
\label{sec-conclusion}

In this work, we present a novel and effective model, ZS-BERT, to tackle the zero-shot relation extraction task. With the multi-task learning structure and the quality of contextual representation learning, ZS-BERT can not only well embed input sentences to the embedding space, but also substantially improve the performance. We have also conducted extensive experiments to study different aspects of ZS-BERT, from hyperparameter sensitivity to case study, and eventually show that ZS-BERT can steadily outperform existing relation extraction models under zero-shot settings. 
Furthermore, learning effective embeddings for relations might also be helpful to semi-supervised learning or few-shot learning by utilizing prototypes of relations as the auxiliary information. 

\section*{Acknowledgements}
This work is supported by Ministry of Science and Technology (MOST) of Taiwan under grants 109-2636-E-006-017 (MOST Young Scholar Fellowship) and 109-2221-E-006-173, and also by Academia Sinica under grant AS-TP-107-M05.

\bibliography{7-Reference}
\bibliographystyle{acl_natbib}


\end{document}